\pdfoutput=1

\documentclass[12pt]{iopart}

\usepackage{amssymb}
\usepackage{hyperref}

\usepackage{graphicx}

\usepackage{cite}
\usepackage{multirow}
\usepackage{setspace}
\usepackage{amsfonts}

\begin{document}

\title{A Systematic Comparison of Supervised Classifiers}

\author{D. R. Amancio$^1$, C. H. Comin$^1$, D. Casanova$^1$, G. Travieso$^1$, O. M. Bruno$^1$, F. A. Rodrigues$^2$ and L. da F. Costa$^1$}
\address{\ \\$^1$Instituto de F\'{\i}sica de S\~{a}o Carlos, Universidade de S\~{a}o Paulo, Av. Trabalhador S\~{a}o Carlense 400, Caixa Postal 369, CEP 13560-970, S\~{a}o
Carlos, S\~ao Paulo, Brazil\\ \ \\}

\address{
$^2$Departamento de Matem\'{a}tica Aplicada e Estat\'{i}stica, Instituto de Ci\^{e}ncias Matem\'{a}ticas e de Computa\c{c}\~{a}o,Universidade de S\~{a}o Paulo, Caixa Postal 668,13560-970 S\~{a}o Carlos,  S\~ao Paulo, Brazil \ \\ \ \\}

\ead{diego.raphael@gmail.com, diego.amancio@usp.br}

\begin{abstract}
Pattern recognition techniques have been employed in a
myriad of industrial, medical, commercial and academic applications. To tackle such a diversity of data, many techniques have been devised. However, despite the long tradition of pattern recognition research, there is no technique that yields the best classification in all scenarios. Therefore, the consideration of as many as possible techniques presents itself as an fundamental practice in applications aiming at high accuracy. Typical works comparing methods either emphasize the performance of a given algorithm in validation tests or systematically compare various algorithms, assuming that the practical use of these methods is done by experts. In many occasions, however, researchers have to deal with their practical classification tasks without an in-depth knowledge about the underlying mechanisms behind parameters. Actually, the adequate choice of classifiers and parameters alike in such practical  circumstances constitutes a long-standing problem and is the subject of the current paper. We carried out a study on the performance of nine well-known classifiers implemented by the Weka framework and compared the dependence of the accuracy with their configuration parameter configurations. The analysis of performance with default parameters revealed that the k-nearest neighbors method exceeds by a large margin the other methods when high dimensional datasets are considered. When other configuration of parameters were allowed, we found  that it is possible to improve the quality of SVM in more than 20\% even if parameters are set randomly.  Taken together, the investigation conducted in this paper suggests that, apart from the SVM implementation, Weka's default configuration of parameters provides an performance close the one achieved with the optimal configuration.

\end{abstract}

\maketitle

\section*{Introduction}

The last decades have been characterized by a progressive increase
of data production and storage.  Indeed, the informatization of most
aspects of human activities, ranging from simple tasks such as phone
calls to shopping habits generates an ever increasing collection of data that can be organized and used for planning.  At the same time, most scientific  projects -- such as in genetics, astronomy and neuroscience -- generate large amounts of data that needs to be analyzed and understood. This trend has given rise to the new term \emph{big data}~\cite{bd1,bd2}. Once such data is organized in a dataset, it is necessary to find patterns concealed in the vast mass of values, which is the objective of \emph{data mining}~\cite{app1,app2,app3,app4,dudahart,bispo,probabilistico}. Because the identification of important patterns (e.g. those that recur frequently or are rare) is impossible to be performed manually, it is necessary to resort to automated pattern recognition.  Nevertheless, it is important to note that pattern recognition remains also critical for organizing and understanding smaller sets of data, such as in medical diagnosis, industrial quality control, and expensive data.

The problem of pattern recognition consists in assigning classes or categories to observations or individuals~\cite{dudahart,bispo,probabilistico}. This can be done in two main ways: (i) with the help of examples or prototypes (\emph{supervised classification}); and (ii) taking into account only the properties of the objects (\emph{unsupervised classification} or \emph{clustering}).  Though seemingly simple, pattern recognition often turns out to be a challenging activity. This is mainly a consequence of \emph{overlap} between the features of different groups in the data, i.e.\ objects in a class have similar properties as those of other classes.  However, several other issues such as choice of features, noise, sampling, also impose further problems while classifying data~\cite{dudahart,bispo,probabilistico}. Even when the features are well-chosen and the data has good quality (properly sampled and without noise), the results of the classification will frequently vary with the choice of different pattern recognition methods.  This situation is typically aggravated for sparse data, presence of noise, or non-discriminative features.   In an attempt to circumvent such problem and to obtain more robust and versatile classifiers, a number of pattern recognition methods have been proposed in the literature~\cite{jainmao,madeira,Smetanin}. Despite the long tradition of pattern recognition research~\cite{dudahart}, there are no definite guidelines for choosing classifiers.  So, those faced with the need to apply pattern recognition are left with the rather difficult task of choosing among several alternative methods.

There are many works in the literature describing which classifiers are more suitable for specific tasks (see e.g.~\cite{some1,some2,some3} and section Related works), but only a few of them consider a systematic quantitative analysis of their performance.
Therefore, in this paper, we assess the performance of the classifiers in carefully chosen datasets, without trying to advocate for any specific method. This means that the dataset used in the study is of fundamental importance to the correct interpretation of the results. Typical datasets employed to compare the performance of different methods include real world and/or artificial data. Advantages of using real datasets include the presence of non-trivial relationships between variables, which may strongly influence the performance of a classifier, the fact that the obtained results will usually be of high confidence when used for samples obtained in the same domain and using a similar criteria, and the presence of noise or unreachable information about the samples (hidden variables). But there is a main drawback associated with using real-world data. Even if one  manage to consistently compare the results obtained with hundreds of real world datasets, the results will still be specific to the datasets being used. Trying to use the information gained in such analyses to a different dataset will most likely be ineffective. Furthermore, obtaining more real data to evaluate other classifier characteristics represents sometimes an arduous task. This is the case of applications whose acquisition process is expensive. For this reason, here we chose  synthetic datasets. Although such datasets are often not representative of specific real-world systems, they can still be used as representations of large classes of data. For example, we can define that all variables in the dataset will have a Pearson correlation of 0.8, and study the behavior of the classifiers when setting this as the main data constrain.

A natural choice of distribution for the variables in the dataset is the multivariate normal distribution. This choice is supported by the well-known central limit theorem \cite{rice_mathematical_2007}, which states that, under certain conditions, the mean of a large number of independent random variables will converge to a normal distribution. This ubiquitous theorem can be used to conclude that, between all infinite possibilities of probability density distributions, the normal distribution is the most likely to represent the data at hand. A second possible choice of data distribution may be the power-law distribution. This is so because there is a version of the central limit theorem stating that the sum of independent random variable with heavy-tailed distributions is generally power-law distributed \cite{willinger_more_2004}. Nevertheless, here we use only normal distribution, leaving power-law distributed variables for a future study.

Since one of our main concerns is making an accessible practical study of the classifiers, we decided to only analyze classifiers available in the Weka software\footnote{Weka is available at \url{http://www.cs.waikato.ac.nz/ml/weka}.}~\cite{witten_data_2011}, which was chosen because of its popularity among researchers. In addition, since the software is open-source, any researcher can look at the code of any specific classifier and confirm the specific procedure being used for the classification. Since Weka has many classifiers available, we decided to select a subset of the most commonly used ones according to~\cite{JainDM00}. 

One distinctive feature of the present work is the procedure we use to compare classifiers. Many works in the literature try to find the best accuracy that a classifier can give and then present this value as the quality of the classifier. The truth is that finding the highest accuracy for a classifier is usually a troublesome task. Additionally, if this high accuracy can only be achieved for very specific values of the classifier parameters, it is likely that for a different dataset the result will be worse, since the parameter was tuned for the specific data analyzed. Therefore, besides giving a high accuracy, it is desirable that the classifier can give such values for accuracy without being too \emph{sensitive} regarding changes of parameters. That is, a good classifier must provide a good classification for a large range of values of its parameters.

In order to study all aforementioned aspects of the classifiers, this work is divided in three main parts. First, we compare the performance of the classifiers when using the default parameters set by Weka. This is probably the most common way researchers use the software. This happens because changing the classifier parameters in order to find the best classification value is a cumbersome task, and many researchers do not want to bother with that. Our second analysis concerns the variation of single parameters of the classifiers, while maintaining other parameters in the default value. That is, we study how the classification results are affected when changing each parameter. Therefore, we look for the parameters that actually matters for the results, and how one can improve their results when dealing with such parameters. Finally, in order to estimate the optimum accuracy of the classifier, as well as verify its sensitivity to simultaneous changes of its parameters, we randomly sample sets of parameter values to be used in the classifier.

The paper is organized as follows. Firstly, we swiftly review some previous works aiming at comparing classifiers. We then describe the generation of synthetic datasets and justify the parameters employed in the algorithm. Next, we introduce the measurements used to quantify the classifiers performance.
We then present a quantitative comparison of classifiers, followed by the conclusions.

\section*{Related works}

Typical works in the literature dealing with comparison between classifiers can be organized into two main groups: (a) comparing among few methods for the purpose of validation and justification of a new approach~\cite{Yang05,TsangKC05,Seetha11,Bezdek86,FanP09}; and (b) systematic qualitative and quantitative comparison between many representative classifiers. Examples of qualitative analysis in (b) can be for example found in ~\cite{WuKQGYMMNLYZS07,JainDM00,Kotsiantis07}. These studies perform a comprehensive analysis of several classifiers, describing the drawbacks and advantages of each method, without considering any quantitative tests. A quantitative analysis of classifiers was performed in~\cite{Demsar06}, where 491 papers comparing quantitatively at least two classification algorithms were analyzed. A comparison of three representative learning methods (Naive Bayes, decision trees and SVM) was conducted in~\cite{HuangL05}, concluding that Naive Bayes is significantly better than decision trees if the area under curve is employed as a performance measurement. Other quantitative studies include the comparison of neural networks with other methods~\cite{Ripley94} and an extensive comparison of a large set of classifiers over many different datasets~\cite{Meyera03}, which showed that SVMs performances very well on classification tasks. Finally,
quantitative comparisons between classifiers can be also found in specific domain problems, such as in Bioinformatics~\cite{Tavares08}, Computer Science~\cite{CufogluLM09,Mico98}, Medicine~\cite{Conrad04,Kuramochi05} and Chemistry~\cite{Berrueta07}.


\section*{Materials and Methods}

In this section we present a generic methodology to construct artificial datasets modeling the different characteristics of real datasets. In addition, we describe the measurements used to evaluate the quality of the classifiers.

\subsection*{Artificial Data}

As noted above, there is a considerable number of reviews in the literature that use real data in order to compare the performance of classifiers. Although this approach is useful when one wants to test the performance for specific classes of data, the small domain of possible cases analyzed renders the results insignificant for a true performance test. Also, with real data it is impossible to systematically study how the classification is being influenced by different variances and correlations between the data, the dimension of the problem, number of classes, distribution of elements per class and, most importantly, the separation between the classes. In order to approach these problems, while having a diversified dataset to test the general purpose classifiers, we use a multivariate Gaussian artificial data generation where many of the parameters chosen are justified by real data, but we can still test variations of them. Data distributions may occur in many different forms. We chose a Gaussian distribution because it has the potential to represent a large ensemble of possible data occurrences on the real world. This observation is supported by the central limit theorem, since it is assumed that the variables are independent and identically distributed \cite{rice_mathematical_2007}.

Here we present a novel method for generating random datasets with a given ensemble of covariances matrices, which was strongly based on the study made by Hirschberger et al. \cite{hirschberger_randomly_2007}. We aim at generating $\mathbb{C}$ classes of data with $\mathbb{F}$ features for each object, with the additional constraint that the number of objects per class is given by the vector $\overrightarrow{N} = (n_1,n_2\ldots n_{\mathbb{C}})$. This problem is mathematically restated as finding $\mathbb{C}$ sets comprising $\mathbb{F}$-dimensional vectors, where each set has a number of elements specified by $\overrightarrow{N}$. $\mathbb{C}$, $\mathbb{F}$ and $\overrightarrow{N}$ are referred to as \emph{strong} parameters, in the sense that they do not bring any information about the relationships between the objects (or vectors). Furthermore, we aimed at generating data complying with the three following constraints:
\begin{itemize}

  \item {\bf Constraint 1}: the variance of the $i$-th feature of each class is drawn from a fixed distribution, $f_{\sigma}$.

  \item {\bf Constraint 2}: the correlation between the $i$-th and $j$-th dimension of each class are drawn from another fixed distribution, $f_{c}$.

  \item {\bf Constraint 3}: we can freely tune the expected separation between the classes, given by parameter $\alpha$, which is explained below.

\end{itemize}
Traditionally, constraints 1 and 2 are not fully satisfied to generate the data. Many studies impose that all the classes display approximately the same variances and correlations, by defining an ensemble of covariance matrices with a fixed spectrum constraint \cite{lin_algorithm_1985,marsaglia_generating_1984}. Unfortunately, this approach is somewhat artificial to generate realistic data, since the assumption that all data classes share similar relationships between their features is quite unlikely. Our approach is more general because, given the shape of the correlation distribution (e.g. U-shaped), the classes can exhibit all kinds of correlations.

\begin{figure*}[!htbp]
	\begin{center}
		\includegraphics[width=0.9\linewidth]{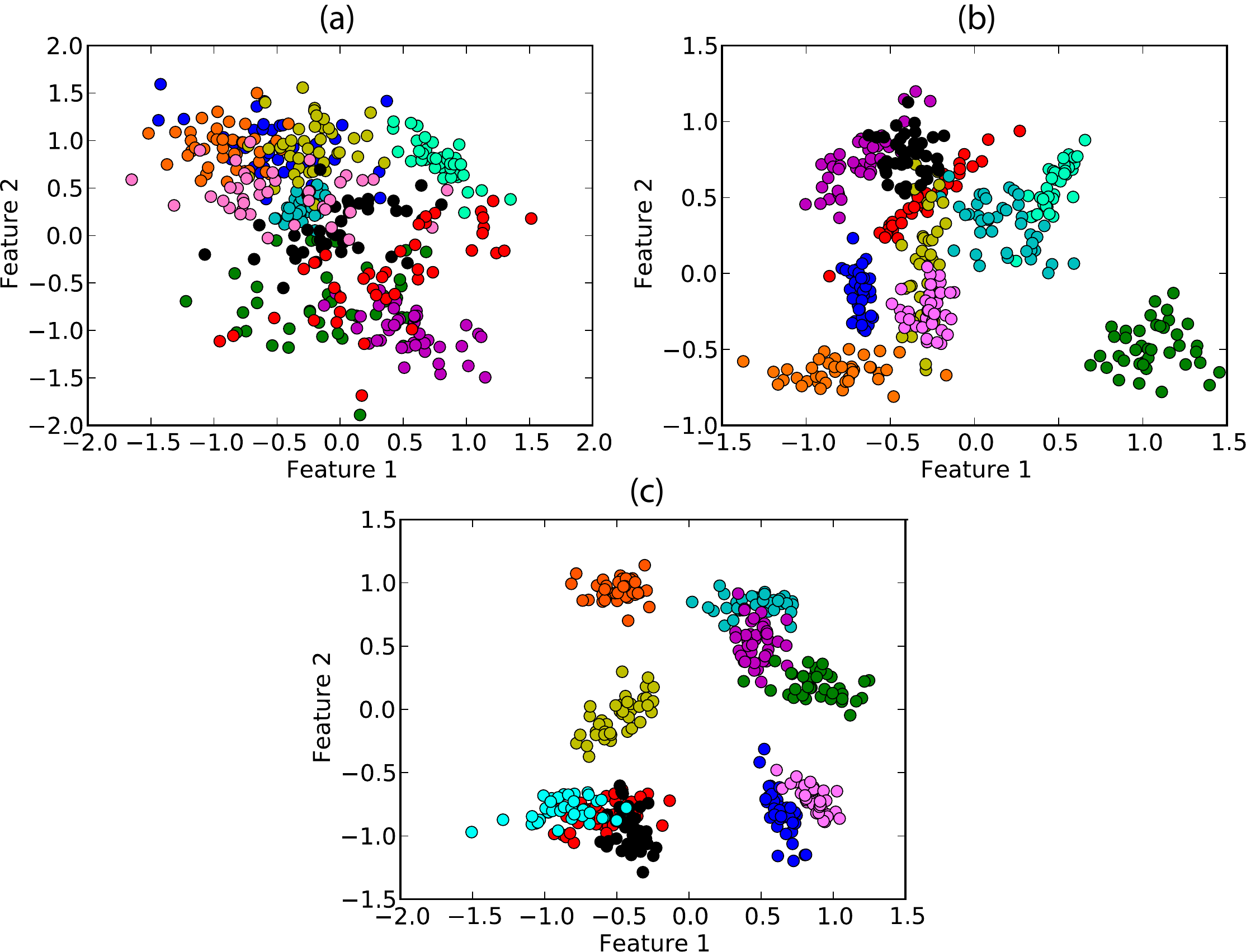}
        \caption{\label{f:random_data}Example of artificial dataset for the case when we have 10 classes and 2 features (DB2F). It is possible to note that different classes have different correlations between the features. The separation between the classes are (a) $\alpha=1$, (b) $\alpha=5$ and (c) $\alpha=7$.}
	\end{center}
\end{figure*}

In order to generate the data with the strong parameters $\mathbb{C}$, $\mathbb{F}$ and $\overrightarrow{N}$ complying with constraints 1, 2 and 3, we need $\mathbb{C}$ covariance matrices (one for each class), where each diagonal and off-diagonal element is drawn, respectively, from $f_{\sigma}$ and $f_{c}$. The most common approach is to randomly draw the mentioned matrix elements from probability density distributions given by $f_{\sigma}$ and $f_{c}$ in order to construct the desired matrices. Unfortunately, this process does not guarantee a valid covariance matrix because every covariance matrix must be positive and semi-definite \cite{horn_matrix_1985}.
%
%
To overcome this problem we use a well-known property stating that for every matrix ${\bf G} \in \mathbb{R}^{n\times m}$, the $n\times n$ matrix ${\bf G}{\bf G}^\textit{T}$ is positive and semi-definite \cite{horn_matrix_1985}. This property allow us to create a random matrix ${\bf G}$ that will generate a valid covariance matrix. The matrix ${\bf G}$ is known as \emph{root} matrix. What is left to us is to define a convenient root matrix so that ${\bf G}{\bf G}^\textit{T}$ follows constraints 1, 2 and 3. Hirschberger et al. \cite{hirschberger_randomly_2007} came up with an elegant demonstration on how to create a covariance matrix following constraints 1 and 2. Actually, by using their methodology it is even possible to set the skewness of $f_c$, but this property will not be employed here, since our goal is to generate off-diagonal elements distributed according to a normal distribution. Using our algorithm, it is possible to create datasets having the following parameters:


\begin{itemize}

  \item {\bf Number of objects per class $\overrightarrow{N}$}: the number of instances in each class can be drawn according to a given distribution. The most common distributions to use are the normal, power-law and exponential distributions. Nevertheless, in order to simplify our analysis, here we use classes having an equal number of instances.

  \item {\bf Number of classes $\mathbb{C}$}: we use $\mathbb{C}=10$. This parameter is not varied throughout the study because we found that the results did not appreciably change for different number of classes (including the binary case $\mathbb{C}=2$).

    \item {\bf Number of features $\mathbb{F}$}: The case $\mathbb{F}=2$ represents the most simple case, since it permits the easy visualization of the data. In order to improve the discriminability of the data, real world datasets oftentimes are described by a larger number of features. Here we vary $\mathbb{F}$ in the range $[2,10]$. Hereafter, we refer to the dataset described by $\mathbb{F}$ features as DB$\mathbb{F}$F.

    \item {\bf Standard deviation of the features}: for each class, the standard deviation of each feature is drawn according to a given distribution $f_{\sigma}$. The process is repeated for each class, using the same distribution $f_{\sigma}$.

    \item {\bf Correlation between features}: for each class, the correlations between the features are drawn according to a given distribution $f_{c}$. The process is repeated for each class using the same distribution. This means that each class of our dataset will show different kinds of correlation. For example, instances from one class may be described by redundant features, while the same features may be much more efficient in describing samples form other classes. The most common choices for $f_{c}$ are: (a) \emph{uniform}, to represent heterogeneous data; (b) Gaussian centered in zero, for mostly uncorrelated; and (c) \emph{U-shaped}, for data with strong correlations. Here we chose a uniform distribution for the correlations.

    \item {\bf Separation between the data ($\alpha$)}: It is a parameter to be varied throughout the experiments, quantifying how well-separated are the classes, compared to their standard deviation. This parameter is simply a scaling of the variance of the features for each class. Since we randomly draw the mean, $m_f$, for each class in the range $-1 \leq m_f \leq 1$, $f_{\sigma}/\alpha$ can be used to define an expected separation between the classes. If $\alpha$ is large, the classes are well-localized and will present little overlap. Otherwise, if $\alpha$ is small, the opposite is true. Clearly, the separation given by $\alpha$ depends on the dimension of the space. Nevertheless, there is no need to define a normalization for $\alpha$, because we are just comparing classifiers and  not different configurations of the data.

\end{itemize}
%
%
Throughout the paper we varied the number of features $\mathbb{F}$ and the separation between classes ($\alpha$). In Figure \ref{f:random_data} we show some examples of the data that can be generated by varying $\alpha$ in a two-dimensional dataset.

\subsection*{Evaluating the performance of the classifiers}

\label{metricas}

A fundamental aspect that should be consider when comparing the performance of classifiers is the proper definition of what \emph{quality} means. It is impossible to define a single metric that will provide a fair comparison in all possible situations. This means that quality is usually specific to the application and, consequently, many measurements have been proposed~\cite{witten:2005}. Nevertheless, there are some measurements that have widespread use in the literature, the most popular being the accuracy rate, f-measure (sometimes together with precision and recall), Kappa statistic, ROC area under curve and the time spent for classification (see~\cite{witten:2005} for a comprehensive explanation of such measurements). Because we are mostly interested in a more practical analysis of the classifiers, we use only the accuracy rate, which is defined as the number of true positives plus the number of true negatives, divided by the total number of instances.

In the literature, oftentimes the average accuracy rate is employed to evaluate the performance of classifiers. This practice is so ubiquitous because many researchers decide to use a number of different kinds of measurements, like the ones previously mentioned, and the specific analysis of each metric turns out to be overly cumbersome. The consequence of such approach is that only the average, and at most the deviation of each metric end up being analyzed. In the present study we only used the accuracy rate. 

To measure the performance of the classifiers, we generate artificial datasets using the method presented in the previous section and calculate some statistics. The principal quantity extracted from each dataset is the average accuracy rate. In addition, we also compute the variation of accuracy across datasets for this quantity is useful to quantify the confidence of the classifier when the dataset is changed. As such, if high values of both average and standard deviation appears then it is possible to state that the classifier performs well, but care must be taken when analyzing a new dataset. The standard deviation of accuracy rate computed over instantiations of the classifier with distinct parameters is useful to quantify the sensitivity with respect to a given parameter.


\section*{Results and Discussion}

The performance of the classifiers was evaluated according to three methodologies. The default values provided by Weka were used in the first strategy.  We then examined the influence of each parameter on the discriminability of the data. Finally, we developed a multivariate strategy. The classifiers considered in the analysis are presented in Table \ref{t:classifier_names}. Throughout the results we used DB2F, DB3F $\ldots$ DB10F to refer to the datasets comprising instances characterized by 2, 3 $\ldots$ 10 features, respectively.

\begin{table}[h]
\caption{\label{t:classifier_names}List of classifiers evaluated in our study. The abbreviated names used for some classifiers are indicated after the respective name.}
\centering
\begin{tabular}{@{}lll}
\hline
{\bf Type} & {\bf Classifier name} & {\bf Name in Weka}\\
\hline
\multirow{2}{*}{Bayesian} & Naive Bayes & bayes.NaiveBayes\\ 
						  & Bayesian Network &  bayes.net (Bayes Net)\\
\hline

\multirow{3}{*}{Tree} & C4.5 & trees.J48 \\ 
				      & Random Forest & trees.RandomForest \\ 
				      & Simple Classification and Regression Tree & trees.SimpleCart \\
\hline
\multirow{1}{*}{Lazy} & k-nearest neighbors (kNN) & lazy.IBk \\
\hline
\multirow{3}{*}{Function} & Logistic & functions.Logistic \\
				      & Multilayer Perceptron & functions.MultilayerPerceptron \\
				      & Support Vector Machine (SVM) & functions.SMO \\
\hline
\end{tabular}
\\ \ \\
\end{table}


\subsection*{Comparison of classifiers using their default parameters}

One of the most typical scenarios concerning practical classification tasks arises when a researcher or practitioner has given a dataset and he/she wants to obtain a good classification without concerning about the parameters of the classifier. Many researchers tackling supervised classification problems do not have an in-depth knowledge of the classifier being employed. As such, all too often they simply compare their results achieved with a few classifiers induced with default parameters and then report the best classification obtained with one of these classifiers. Therefore, comparing accuracy rates obtained with the default parameters of the classifiers are of fundamental importance because this case represents the most common use case.

\begin{figure*}[h]
	\begin{center}
    \includegraphics[width=0.75\linewidth]{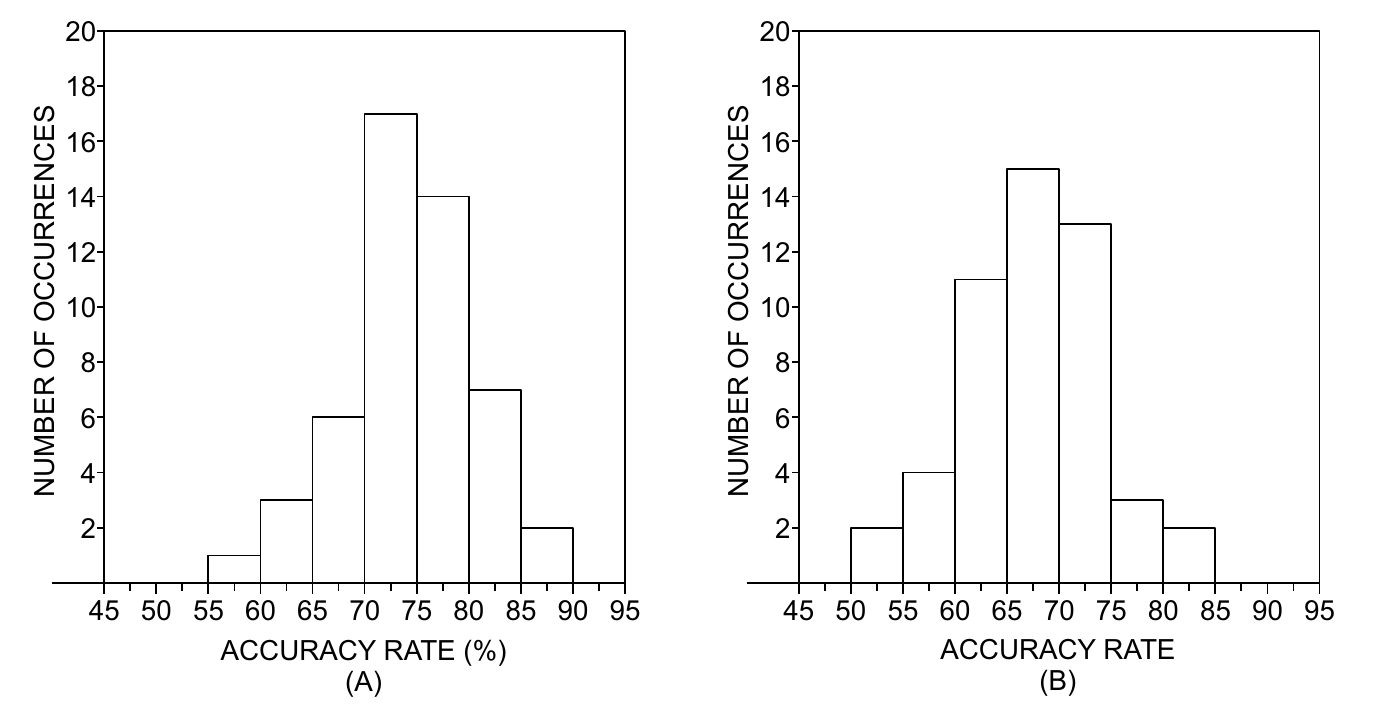}
        \caption{Histogram displaying the distribution of accuracy rates achieved with default parameters for (a) Naive Bayes; and (b) SVM; in DB2F.}
        \label{f:example_default}
	\end{center}
\end{figure*}

The performance of each classifier to classify instances in DB2F considering Weka's default parameters\footnote{Section 1 of the Supporting Information (SI) provides a brief description of classifiers and parameters.}\footnote{The Supporting Information is available from \url{https://dl.dropboxusercontent.com/u/2740286/siComparisons.pdf}.} is summarized in Table~\ref{t:default_two}. {This table provides sufficient information to describe the performance of the classifiers for the default parameters, since we observe that the distribution of accuracy values across all datasets is a Gaussian function}.
Figure \ref{f:example_default} shows the distribution of accuracy rates obtained with Naive Bayes and SVM in DB2F. All classifiers performed well, as revealed by high values of average accuracy rates. Surprisingly, the Naive Bayes classifier, which is based on a naive assumption of feature independence, outperformed cutting-edge classifiers. Even if we consider the significance of the differences we found that Naive Bayes performs significantly better than C4.5, Simple Cart, Bayes Net, Random Forest, kNN and SVM (see Table S1 of the Supporting Information).
Another surprising result refers to the worst ranking obtained by the SVM classifier, which recently have drawn much attention owing to recent results reporting that this classifier usually outperforms other traditional pattern recognition techniques~\cite{Meyera03}. The variance of accuracy reveals that all the classifiers give similar confidence for use across different datasets. The best and worst cases tend to follow the same trend observed for the average accuracy, with the Naive Bayes giving the best result overall and the kNN displaying the worst classification for a particular dataset. It is noteworthy that, even though we have employed the same parameters to create the 50 datasets in DB2F, the difference between the best and worst cases is considerably high, being as large as 30\% in some classifiers (see e.g. the C4.5 classifier).

\begin{table}[h]
\caption{\label{t:default_two}Simple statistics summarizing the accuracy rate obtained in DB2F (dataset with two features) when all the parameters are set with their respective default values.}
\centering
\begin{tabular}{@{}llcccc}
\hline
\multirow{2}{*}{\#} & \multirow{2}{*}{\bf Classifier} & {\bf Average} & {\bf Deviation} & {\bf Best case} & {\bf Worst case} \\
&  & (\%) & (\%) & (\%) & (\%)\\
\hline
\textbf{1} & Naive Bayes & 74.39 & 6.61 & 88.50 & 59.00 \\
\textbf{2} & Logistic & 72.67 & 6.67 & 86.25 & 58.75 \\
\textbf{3} & Perceptron & 72.67 & 6.38 & 85.50 & 59.50 \\
\textbf{4} & C4.5 & 70.24 & 6.66 & 84.75 & 54.50 \\
\textbf{5} & Simple Cart & 70.23 & 6.48 & 86.25 & 57.00 \\
\textbf{6} & Bayes Net & 68.68 & 7.00 & 85.25 & 54.75 \\
\textbf{7} & Random Forest & 69.94 & 6.80 & 83.25 & 53.75 \\
\textbf{8} & kNN & 68.67 & 7.49 & 82.50 & 51.00 \\
\textbf{9} & SVM & 67.44 & 6.62 & 82.75 & 52.00 \\
\hline
\end{tabular}
\\ \ \\
\end{table}

%
%

An important question that can be posed refers to the dependency of the ranking observed in Table \ref{t:default_two} with the number of features describing the instances. To address this issue, we evaluated the quality of classifiers in DB10F (i.e., an dataset described by a set of 10 features). Table \ref{t:default_ten} gives the statistics concerning the accuracy of the classifiers evaluated DB10F. It is apparent that the difference in average accuracy is now much higher compared to the differences obtained in DB2F. Surprisingly, the kNN classifier turned out to provide, by a large margin, the highest average accuracy. This result is particularly impressive if we perceive that the usual number of neighbors analyzed (i.e. the parameter $K$) to infer the class of new instances is $K=1$.
Another interesting difference arising from the comparison of the results obtained with DB2F (Table \ref{t:default_two}) and DB10F (Table \ref{t:default_ten}) is that the standard deviation take lower values in the latter. This means that the statistical difference between classifiers is much higher for ten features.
The best and worst values follow the trend observed for the average, with the surprising result that the worst case for kNN is better than almost \emph{all} values obtained for the other classifiers, with the only exception being the Perceptron.

\begin{table}
\centering
\caption{\label{t:default_ten}Simple statistics regarding the accuracy rate obtained in DB10F when all the parameters of the classifiers are set with their respective default values.}
\begin{tabular}{@{}llcccc}
\hline
\multirow{2}{*}{\#} & \multirow{2}{*}{\bf Classifier} & {\bf Average} & {\bf Deviation} & {\bf Best case} & {\bf Worst case} \\
 &  & (\%) & (\%) & (\%) & (\%)\\
\hline
\textbf{1} & kNN & 94.28	& 1.76 & 97.50 & 90.00 \\
\textbf{2} & Perceptron & 83.65	& 3.94	& 91.75	& 74.00 \\
\textbf{3} & Random Forest &  80.14	& 2.83 & 86.00 & 71.50 \\
\textbf{4} & Naive Bayes & 76.78	& 4.23 & 85.00 & 60.25  \\
\textbf{5} & SVM & 74.01	& 4.70 & 84.25 & 59.75  \\
\textbf{6} & Logistic &  71.16 & 4.69 & 80.25 & 59.00 \\
\textbf{7} & Simple Cart & 71.07	& 4.71 & 80.25 & 59.00 \\
\textbf{8} & C4.5 & 65.70 & 3.62 & 73.75 & 56.75 \\
\textbf{9} & Bayes Net & 56.87 & 5.16 & 67.00 & 41.25 \\
\hline
\end{tabular}
\\ \ \\
\end{table}

\begin{figure*}[!htbp]
\begin{center} \includegraphics[width=0.85\linewidth]{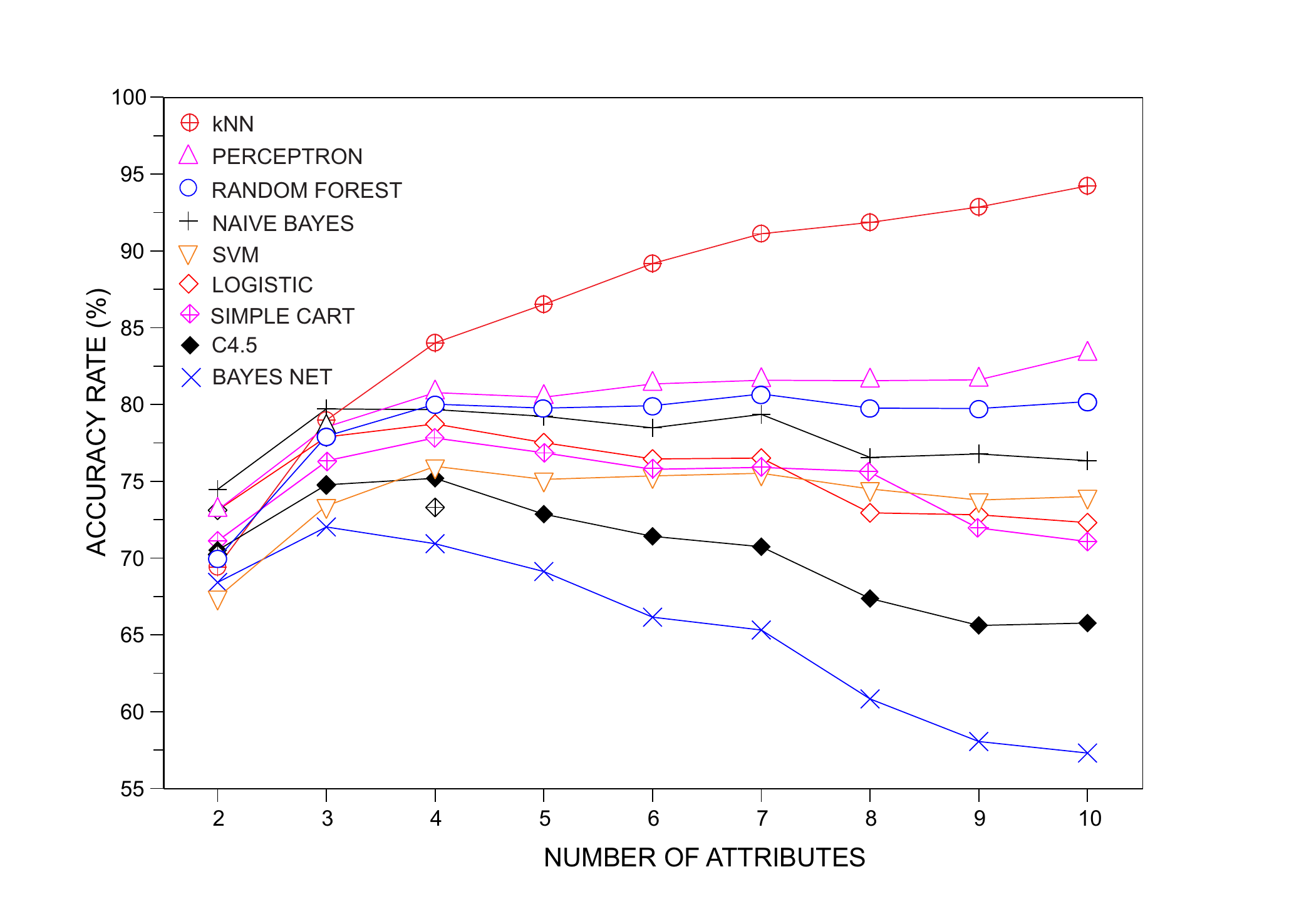}
        \caption{\label{f:vary_num_features}Behavior of the accuracy rates as the number of attributes employed to describe the instances is increased. As more attributes are taken into account, the kNN becomes significantly better than other pattern recognition techniques.}
	\end{center}
\end{figure*}

Another interesting issue to be investigated concerns the assessment of the discriminability of the classifiers as the number of features varies continually from two to ten features. To perform this analysis, we assessed the performance of the classifiers in DB2F, DB3F $\ldots$ DB10F. In Figure \ref{f:vary_num_features} we show the variation of the average accuracy as the number of features describing the dataset is incremented. Three distinct behavior are evident: (i) the accuracy increases; (ii) the accuracy is nearly constant; and (iii) the accuracy decreases. The only classifier in which the pattern (i) was observed was the kNN classifier. The behavior (ii) is the most common trend. Finally, the third effect is more prominent in the case of the C4.5, Simple Cart and Bayes Net. All in all, these results suggest that in high-dimensional classifications tasks the kNN performs significantly better than others classifiers. Therefore, the proper choice of the classifier is critical to yield high-quality classifications in unseen instances (note that the difference between kNN and Bayes Net in DB10F is almost 40\%).

\subsection*{Varying parameters: one-dimensional analysis}

\begin{figure*}[!htbp]
	\begin{center}
		\includegraphics[width=1\linewidth]{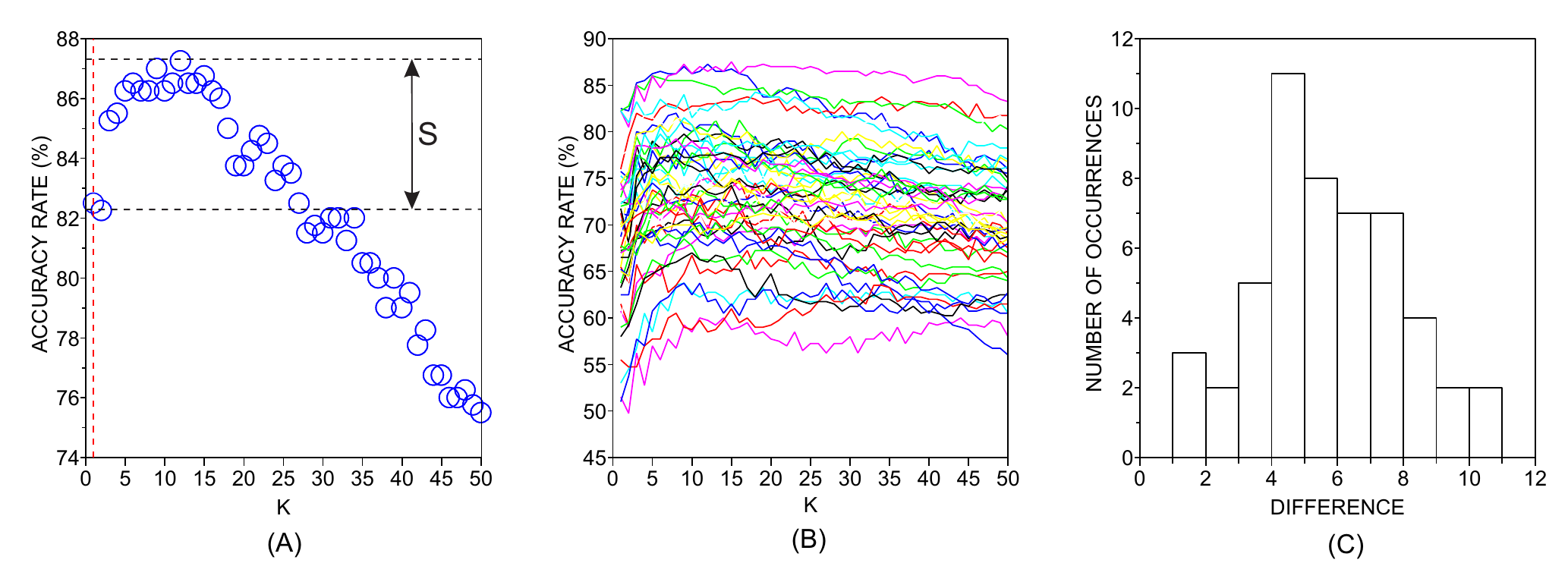}
        \caption{\label{t:vary_one_parameter}One dimensional analysis performed with the parameter $K$ of the kNN classifier. Panel (a) illustrates the default value of the parameter ($K=1$) with a red vertical dashed line. The accuracy rate associated with default values of parameters is denoted by $\Gamma_{def}$ and the best accuracy rate observed in the neighborhood of the default value of $k$ is represent as $\Gamma_{max}$. The difference between these two quantities is represented by $S(K) = \Gamma_{max} - \Gamma_{def}$. Panel (b) shows how the accuracy rates varies with the variation of $K$ in DB2F (each line represent the behavior of a particular dataset in DB2F). Finally, panel (c) displays the distribution of $S(K) = \Gamma_{max} - \Gamma_{def}$ in DB2F.}
	\end{center}
\end{figure*}

An alternative scenario in typical classification tasks arises when the researcher or practitioner wants to improve the performance of the classification by setting up the suitable values of the parameters. In this case, we turn to the concept of \emph{sensitivity} of the classification with regard to the parameters. In other words, if a good classification is achieved only for a very small range in the parameter space, then for many applications it will be very difficult to achieve the best accuracy rate provided by the classifier. Conversely, if the  classifier provides high accuracy rates for many different configuration of parameters, then one expects that it will consistently yield high-quality classifications for the researcher or practitioner who aims at obtaining the best accuracy rate with a minimal effort spent tuning the suitable values of the parameters.
To probe the sensitivity of the classifiers with regard to distinct values or parameters, we analyzed the behavior of the accuracy rates curves when each parameter is varied separately while retaining the remaining parameters set at their default values. This one-dimensional analysis is illustrated in Figure~\ref{t:vary_one_parameter}. The behavior of the accuracy with the adoption of values different from the default for some parameters is shown in Figures~S1 and S2 of the SI.

An extensive analysis comparing the quality achieved with the parameters set with default and non-default values is provided in Table~\ref{tab:comparacao1} for classifications obtained in DB2F. For each parameter, we provide the average $\langle S \rangle$, the standard deviation $\Delta S$ and the maximum value of $S(p) = \Gamma_{max} - \Gamma_{def}$, where $\Gamma_{max}$ and $\Gamma_{def}$ are respectively the maximum accuracy observed when the parameter $p$ varies and the accuracy rate obtained with all the parameters set at their default values. Therefore, the statistic computed over $S(p)$ quantifies how much the accuracy of the classification is influenced when each parameter is set to a value different from the default. Table \ref{tab:comparacao1} shows that $\langle S \rangle \leq 0$ for almost all parameters, with the only exceptions being the number of seeds ($K$) of the kNN and the type ($S$) of Support Vector Machine used. This result suggests that the default parameters usually provide a classification performance that is close to the optimum. Interestingly, even the maximum gain in accuracy is usually small, as they do not exceed~6.25\% in any particular dataset (aside from the kNN and SVM classifiers).

\begin{table}[htbp]
\centering
\small
\caption{\label{tab:comparacao1}Comparison between the accuracy achieved with the default and the best parameter. The difference between the former and the latter in in DB2F was summarized with the average, the standard deviation and the maximum difference.}
		\begin{tabular}{@{}lcccc}
			\hline
\multirow{2}{*}{\bf Classifier} & \multirow{2}{*}{\bf Parameter} & $\langle S \rangle$ & $\Delta S$ & $\max S$ \\
 & & (\%) & (\%) & (\%) \\
            \hline
            {Bayes Net} & -D &	0.00	 &	0.00	  &  0.00 \\
            kNN         & -K &  6.62     &	2.45  	  & 12.75 \\
            kNN         & -I &  0.00     &  0.00      &	0.00 \\
            kNN         & -F &  0.00	 &  0.00  	  & 0.00 \\
		    kNN         & -X &  0.00	 &  0.00  	  & 0.00 \\
            C4.5         & -U &	-0.18    &  0.72 	  &  1.25 \\
            C4.5         & -S &  0.04	 &  0.26	  &  1.00 \\		 C4.5         & -A &	 0.00	 &	0.00      &  0.00 \\
            C4.5         & -C &	0.69	 &  0.54	  &  2.00 \\
            C4.5         & -M &	0.86	 &  0.76  	  &  2.75 \\
            C4.5         & -N &	0.23	 &  1.36  	  &  2.75 \\
            Logistic    & -R &	0.63     &	0.60  	  &  2.25 \\
            Logistic    & -M &	0.84     &	0.61	  &  2.75 \\
            Naïve Bayes & -K &	-0.74	 &  1.15  	  &  1.75 \\
            Naïve Bayes & -D & 	-5.79  	 &  3.64  	  &  1.25 \\
            Perceptron  & -D &	-51.25	 &  7.17  	  & -37.75 \\
            Perceptron  & -C &	0.00	 &	0.00	  &  0.00 \\
            Perceptron  & -H & 	1.74	 &  1.61      &	 6.25 \\
            Perceptron  & -L &	1.30	 &  0.88   	  &  3.75 \\
            Perceptron  & -M &	1.17  	 &  0.83 	  &  3.75 \\
            Perceptron  & -N & 	1.00	 &	0.66 	  &  3.00 \\
            Perceptron  & -V &	0.74	 &  0.75      &  2.50 \\
            Perceptron  & -E &	0.00	 &	0.00	  &	 0.00 \\
            Random Forest &-I & 0.02     &  0.14      &  1.00 \\
            Random Forest &-K &-0.09     &  0.64      &  -4.50 \\
            Random Forest &-depth& 0.03  &  0.18      &  1.25 \\
            Random Forest & -S   & 0.04  &  0.28      &  2.00 \\
            Simple Cart   & -S   & 0.06  &  0.39      &  2.75 \\
            Simple Cart   & -C   & 0.00  &  0.00      &  0.00 \\
            Simple Cart   & -M   & 0.04  &  0.25      &  1.75 \\
            Simple Cart   & -N   & 0.02  &  0.11      &  0.75 \\
            Simple Cart   & -A   & 0.01  &  0.07      &  0.50 \\
            Simple Cart   & -H   & 0.00  &  0.00      &  0.00 \\
            Simple Cart   & -U   & -0.01 &  0.07      &  -0.5 \\
            SVM           & -C   & 0.05  &  0.32      &  2.25 \\
            SVM           & -L   & 0.01  &  0.07      &  0.50 \\
            SVM           & -P   & 0.03  &  0.21      &  1.50 \\
            SVM           & -V   & 0.00  &  0.00      &  0.00 \\
            SVM           & -N   & 0.03  &  0.21      &  1.50 \\
            SVM (poly kernel) & -E & 1.38 &  1.29      &  4.50 \\
            SVM (NP kernel)   & -E & -20.87& 5.28 & -8.00 \\
            SVM (RBF kernel)     & -G & 2.55 &  2.55      &  12.75 \\
            SVM (Puk kernel)     & -S & 5.88    & 2.46       & 11.75 \\
			\hline
		\end{tabular}
\\ \ \\
\end{table}

Similarly to Table~\ref{tab:comparacao1}, Table~\ref{tab:comparacao2} shows the results for single parameter variation for classifications performed in DB10. We note that a proper analysis of this table must
consider the  accuracy rate obtained with default parameters (see Table~\ref{t:default_ten}), because the latter has a large variation across classifiers. Therefore, if a classifier performs very well with parameters set with their default the values, one expects that a significant improvement through one-dimensional variation of parameters will be less probable. This effect becomes evident when one analyze for example the kNN. The default configuration of parameters yields high accuracy rates ($\Gamma_{def} \simeq 94\%$), while the average improvement through one-dimensional analysis is only $\Delta S(K) = 0.01\%$. A significant improvement in the discriminability was observed for the Perceptron through the variation of the size of the hidden layers (H). In a similar manner, a significant increase of accuracy was observed when we varied the number of trees (I) of the Random Forest.
As for the SVM classifier, six of its parameters allowed an increase of about 20\%, which led to accuracy rates higher than 94\% in many cases. This result suggest that the appropriate parameter tuning in SVM might improve significantly its discriminability.

\begin{table}[htbp]
\small
	\centering
    	\caption{\label{tab:comparacao2}Comparison between the accuracy achieved with the default and the best parameter. The difference between the former and the latter in DB10F was summarized with the average, the standard deviation and the maximum difference.}
		\begin{tabular}{@{}lcccc}
			\hline
\multirow{2}{*}{\bf Classifier} & \multirow{2}{*}{\bf Parameter} & $\langle S \rangle$ & $\Delta S$ & $\max S$ \\
 & & (\%) & (\%) & (\%) \\
            \hline
       		{Bayes Net} & -D & 0.00 & 0.00 & 0.00 \\
            kNN         & -K &  0.01 & 0.04 & 0.25 \\
            kNN         & -I &   0.00 & 0.00 & 0.00 \\
            kNN         & -F &   0.00 & 0.00 & 0.00  \\
		    kNN         & -X &   0.00 & 0.00 & 0.00 \\
            C4.5         & -U & -0.05 & 0.29 & 0.75 \\
            C4.5         & -S &  -0.01 & 0.13 & 0.25 \\		
            C4.5         & -A &	  0.00 & 0.00 & 0.00 \\
            C4.5         & -C &	 0.27 & 0.30 & 1.25 \\
            C4.5         & -M &	 1.32 & 0.96 & 3.50 \\
            C4.5         & -N & -7.44 & 1.75 & -2.75 \\
            Logistic    & -R &   0.58 & 0.71 & 4.25  \\
            Logistic    & -M &	 0.81 & 0.73 & 4.25 \\
            Naive Bayes & -K &	-2.91 & 1.64 & 1.25 \\
            Naive Bayes & -D &  -19.20 & 3.10 & -11.75 \\
            Perceptron  & -D &-56.10 & 5.33 & -46.50 \\
            Perceptron  & -C &	  0.00 & 0.00 & 0.00 \\
            Perceptron  & -H &  7.06 & 2.53 & 13.25 \\
            Perceptron  & -L &	2.27 & 1.11 & 5.50 \\
            Perceptron  & -M &	 2.33 & 1.00 & 4.25  \\
            Perceptron  & -N &   1.01 & 0.77 & 4.00 \\
            Perceptron  & -V &  0.45 & 0.76 & 2.75 \\
            Perceptron  & -E &	  0.00 & 0.00 & 0.00 \\
            Random Forest &-I & 5.67 & 1.73 & 10.50      \\
            Random Forest &-K & 0.54 & 0.98 & 3.75      \\
            Random Forest &-depth& 1.11 & 1.03 & 3.75   \\
            Random Forest & -S & 3.04 & 1.86 & 8.75     \\
            Simple Cart   & -S   & 1.09 & 0.81  & 2.75  \\
            Simple Cart   & -C   & 0.00 & 0.00  & 0.00  \\
            Simple Cart   & -M   & 1.41 & 1.16  & 4.25  \\
            Simple Cart   & -N   & 1.31 & 0.92  & 3.25  \\
            Simple Cart   & -A   & 3.89 & 1.70  & 9.00  \\
            Simple Cart   & -H   & 0.00 & 0.00  & 0.00  \\
            Simple Cart   & -U   & - 1.21 & 1.18 & -4.00 \\
            SVM           & -C   & 22.15  & 4.10 & 36.50  \\
            SVM           & -W   & 0.38 & 0.33 & 1.50 \\
            SVM           & -P   & 0.52 & 0.67 & 3.25 \\
            SVM           & -V   & 0.00 & 0.00 & 0.00 \\
            SVM           & -N   & 23.54 & 4.60 & 39.30 \\
            SVM (poly kernel) & -E & 23.57 & 4.40 & 37.75   \\
            SVM (NP kernel)   & -E & 19.08 & 4.84  & 31.00 \\
            SVM (RBF kernel)     & -G & 21.55 & 3.91  & 34.75 \\
            SVM (Puk kernel)     & -S & 19.90 & 3.59  & 32.50 \\
			\hline
		\end{tabular}
\\ \ \\
\end{table}

\subsection*{Multidimensional analysis}

\begin{figure*}[!htbp]
	\begin{center}
	\includegraphics[width=0.75\linewidth]{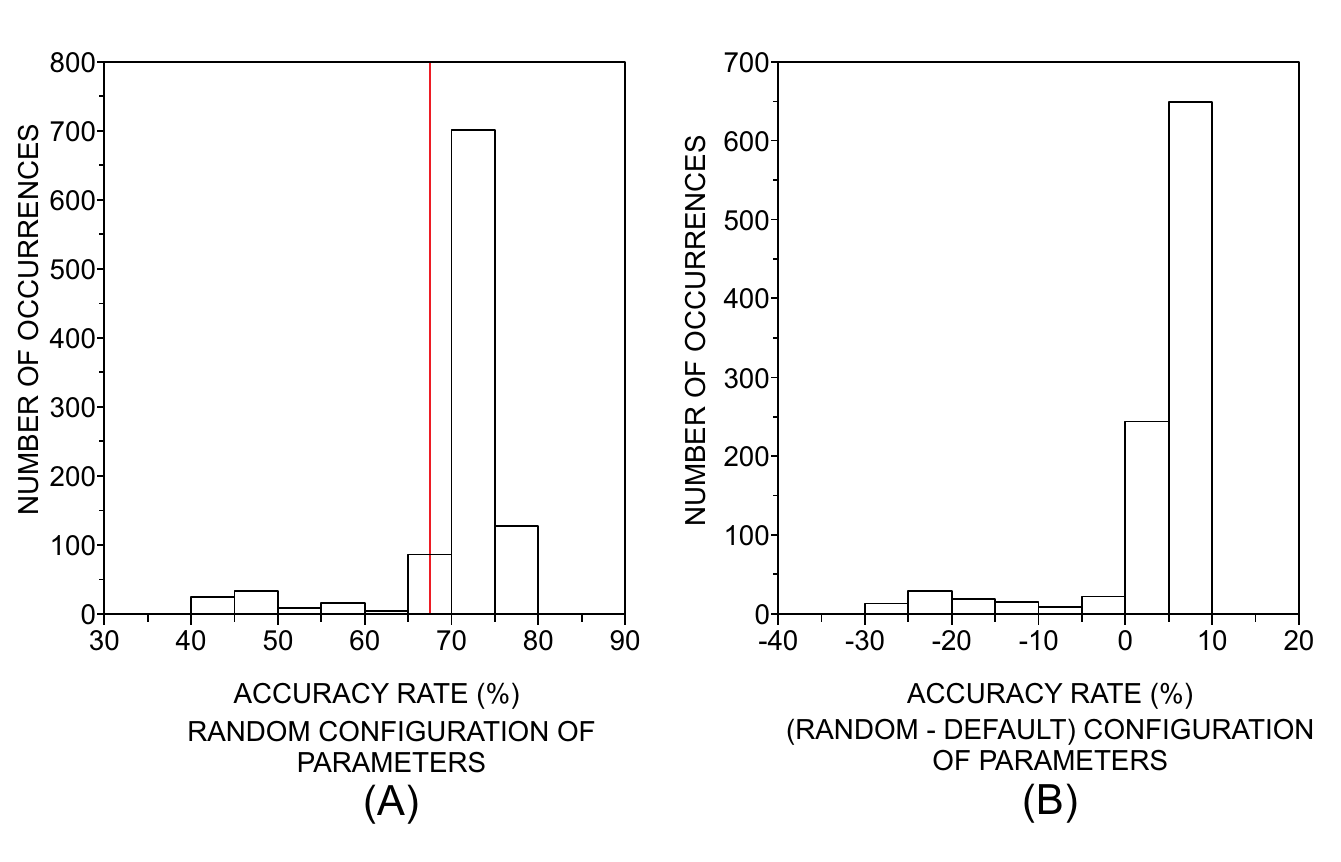}
    \caption{\label{f:exp_random_parameters}Example of the random parameters analysis. We use one of the artificial datasets and the kNN classifier. (a) By randomly drawing 1,000 different parameter combinations of kNN we construct an histogram of accuracy rates. The red dashed line indicates the performance achieved with default parameters. (b) The default value is subtracted from the values obtained for the random drawing. The normalized area of the histogram for values that are above zero indicates how easy is to improve the performance with a  random tuning of parameters.}
	\end{center}
\end{figure*}

Although the one-dimensional analysis is useful to provide relevant information regarding the variability of accuracy  with regard to a given parameter, this type of analysis deliberately disregards the influence of possible mutual interdependencies among parameters on the performance of the classifiers. In order to consider this interdependence, we randomly sample the values of parameters in a bounded range. More specifically, $1,000$ random configurations of parameters for each classifier was generated and each classifier was applied to discriminate the classes in DB2F and DB10F\footnote{The Naive Bayes and Bayesian Net classifiers were not included in the multidimensional analysis, since they only have binary parameters}. We then compared the performance of the best random configuration with the performance achieved with the default parameters. An example of the procedures adopted in the multidimensional analysis is provide in Figure~\ref{f:exp_random_parameters}. A more `efficient' possibility could be based on the search of the best accuracy rates (considering all configuration of parameters) through an optimization heuristic. Nevertheless, we decided not to employ optimization heuristics because the search process would present itself many problems caused by the different kinds of parameters used by distinct classifiers (e.g., nominal, binary, integer, etc). Moreover, we would have to overcome similar parameter optimization when setting up the parameters of the optimization heuristics.

\begin{figure*}[!htbp]
	\begin{center}
	 \includegraphics[width=0.9\linewidth]{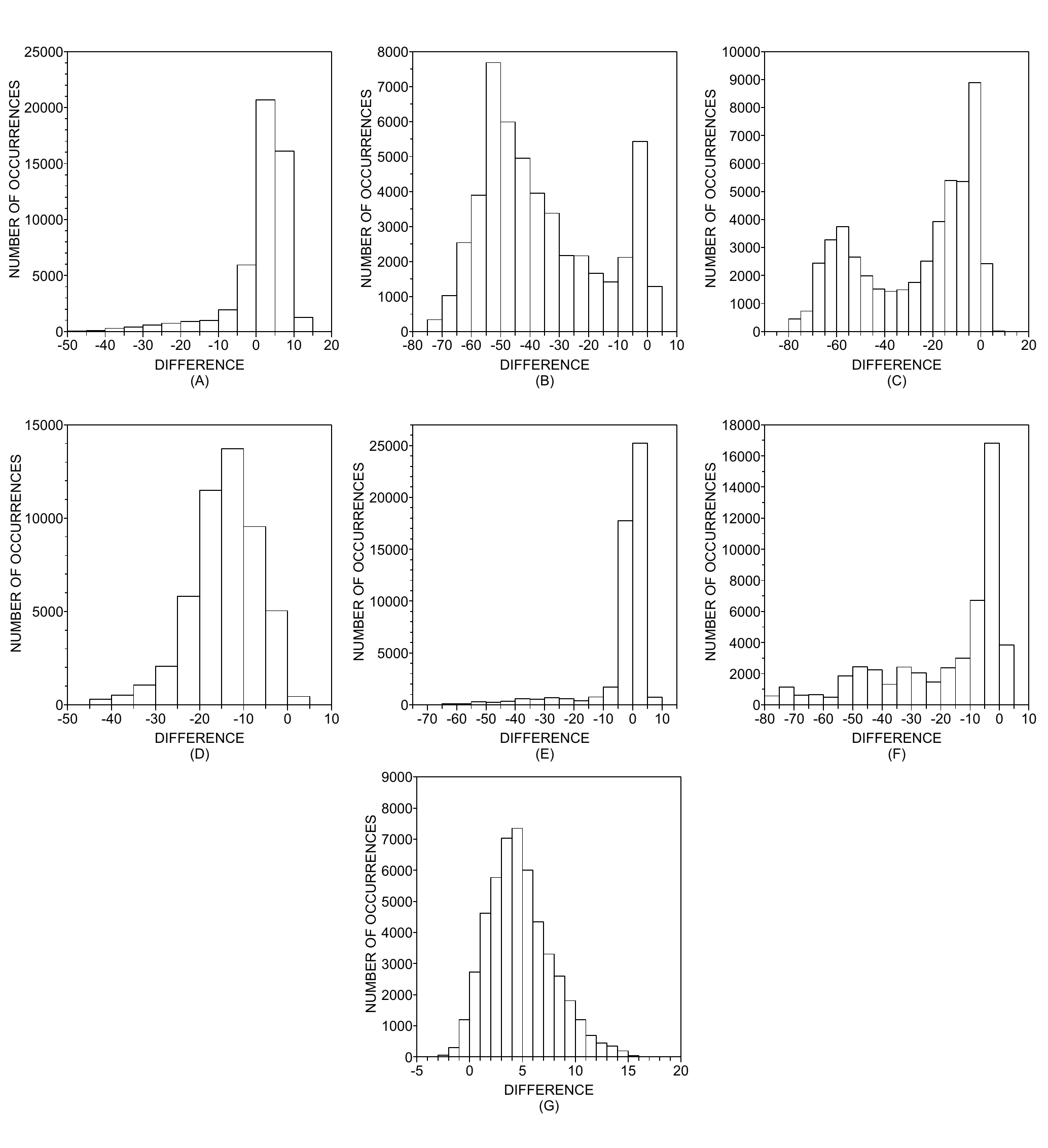}
        \caption{Distribution of the difference of accuracy rates observed between the random and default configuration of parameters. (a) kNN; (b) C4.5; (c) Perceptron; (d) Logistic; (e) Random Forest; (f) Simple Cart; (g)SVM.
        Note that, in the case of kNN and SVM classifiers, most of the random configurations yields better results than the default case. Differently, the same distribution for the C4.5, Perceptron, Logistic, Random Forest and Simple Cart reveals that the default configuration yields accuracy rates similar to the default case. }~\label{f:random_parameter1}
	\end{center}
\end{figure*}

In Figure~\ref{f:random_parameter1}, we show the histograms of the accuracy rates obtained with the random choice of parameters of the classifiers, which were evaluated in DB2F. In order to summarize the main characteristics observed in these histograms, in Table \ref{tab:multi1} we show some statistics taken over the histograms. The $p$-value quantifies the percentage of realizations in which a random configuration of parameters outperformed the performance obtained with the default configuration. Considering these cases where an improvement is observed, we can summarize the values of accuracy by taking the average, standard deviation and maximum value. It is noteworthy that the random choice of parameters usually reduces the accuracy (i.e. $p$-value $< 50.0\%$) for Simple Cart, Perceptron, C4.5 and Logistic. This means that one should be aware when choosing parameters other than the default configuration, since most of the random configurations impacts the performance negatively. Surprisingly, in almost every random choice of parameters (96.89\% of the cases) the accuracy of the SVM increases. In the case of the kNN, the improvement is less likely ($p$-value = 76.15\%). The Random Forest shows a typical small improvement in 52\% of the realizations, in comparison with SVM and kNN. Whenever the computing time for each dataset is not very high, it is possible to generate many random configurations and select the one providing the highest accuracy. In this case, the most important parameter extracted from Table~\ref{tab:multi1} becomes the maximum accuracy. This scenario is particularly useful for SVM, kNN and Random Forest, since the performance can be improved in 16\%, 13\% and 10\%, respectively. Actually, SVM an kNN emerge as the best classifiers when we consider only the best realization among the 1,000 random configurations for each dataset (see Table~\ref{nova}).

\begin{table}[htbp]
	\centering
    \caption{\label{tab:multi1}Comparison of classifiers using a multidimensional analysis of classifiers evaluated in DB2F. $p$-value represents the percentage of cases where the random configuration of parameters yields a classifiers that outperforms the classifier obtained with default parameters. Mean, deviation and maximum refers
    to the increase in accuracy provided by the random configuration, when it outperforms the default configuration. In the case of the SVM, the random choice of parameters yields a classification more accurate than the default classification in about 97\% of the cases. In this case, the average and maximum improvement of quality are 5\% and 16\%, respectively.}
		\begin{tabular}{@{}llcccc}
			\hline
			\multirow{2}{*}{\#} & \multirow{2}{*}{\bf Classifier} & \multirow{2}{*}{\bf $p$-value} & {\bf Mean} & {\bf Deviation} & {\bf Maximum} \\
 &  &  & (\%) & (\%) & (\%) \\
            \hline
            {\bf 1} & SVM & 96.89 & 5.07 & 2.87 & 16.00 \\
            {\bf 2} & kNN & 76.15 & 4.90 & 2.54 & 13.00 \\
            {\bf 3} & RandomForest & 51.93 & 1.82 & 1.34 & 10.25 \\
            {\bf 4} & Simple Cart & 7.68 & 0.91 & 0.57 &  3.75 \\
            {\bf 5} & Perceptron & 4.87 & 1.28 & 1.00 & 6.75 \\
            {\bf 6} & C4.5 & 2.56 & 0.93 & 0.69 & 3.50 \\
			{\bf 7} & Logistic & 0.88 & 0.77 & 0.48  & 2.75 \\
        \hline
		\end{tabular}
\\ \ \\

\end{table}

\begin{table}[h]
\caption{\label{nova}Ranking of classifiers in DB2F considering the best configuration of parameters among the 1,000 random configurations.}
\centering
\begin{tabular}{@{}llcc}
\hline
\multirow{2}{*}{\#} & \multirow{2}{*}{\bf Classifier} & {\bf Average} & {\bf Deviation}  \\
&  & (\%) & (\%) \\
\hline
\textbf{1} & SVM & 78.1 & 5.0  \\
\textbf{2} & kNN  & 75.9 & 6.2  \\
\textbf{3} & Perceptron & 75.4 & 6.2  \\
\textbf{4} & Random Forest & 77.3 & 5.1  \\
\textbf{5} & Logistic & 73.5 & 6.0  \\
\textbf{6} & Simple Cart & 72.8 & 6.3 \\
\textbf{7} & C4.5 & 71.5 & 6.5  \\
\hline
\end{tabular}
\\ \ \\
\end{table}


Repeating the above analysis for the classifications performed in DB10F, one observe some differences in the results, which are shown in Table \ref{tab:multi2}. From the analysis of the means (third column), it is clear that, apart from SVM, a significant improvement in accuracy is much less likely. This results reinforce the premise that default parameters already provide an  accuracy that is near to the optimum. As we found for DB2F, the performance of SVM can be significantly improved by the suitable configuration of parameters. Note that the average improvement of 20.35\% is equivalent to the one found with an uni-dimensional variation in the complexity parameter (see parameter C in Table \ref{tab:comparacao2}). Therefore, to get the best configuration of the SVM we need to vary only one parameter. Again, if we consider only the best configuration among the 1,000 random configurations for each dataset, the SVM and kNN performs better than the other methods (see Table \ref{nova2}).

\begin{table}[htbp]
	\centering
    \caption{\label{tab:multi2}Comparison of classifiers using a multidimensional analysis of classifiers evaluated in DB10F. $p$-value represents the percentage of cases where the random configuration of parameters yields a classification that outperforms the classification obtained with default parameters. Mean, deviation and maximum refers
    to the increase in accuracy provided by the random configuration, when it outperforms the default configuration. In the case of the SVM, the random choice of parameters yields a classification more accurate than the default classification in about 99\% of the cases. When this scenario occurs, the average and maximum improvement of quality are 20\% and 39\%, respectively.}
		\begin{tabular}{@{}llcccc}
			\hline
			\multirow{2}{*}{\#} & \multirow{2}{*}{\bf Classifier} & \multirow{2}{*}{\bf $p$-value} & {\bf Mean} & {\bf Deviation} & {\bf Maximum} \\
 &  &  & (\%) & (\%) & (\%) \\
            \hline
            {\bf 1} & SVM & 99.43 & 20.35 & 5.67 & 39.00 \\
            {\bf 3} & Random Forest & 48.74 & 3.91 & 2.25 & 14.5 \\
            {\bf 2} & kNN & 21.84 & 0.29 & 0.10 & 0.75 \\
            {\bf 4} & Simple Cart & 4.95 & 1.89 & 1.28 & 7.25 \\
            {\bf 5} & Perceptron & 4.11 & 3.25 & 2.46 & 12.00 \\
            {\bf 7} & Logistic & 1.27 & 0.76 & 0.48 & 3.75 \\
            {\bf 6} & C4.5 & 0.47 & 1.23 & 0.90 & 3.50 \\
            \hline
		\end{tabular}
\\ \ \\
\end{table}

\begin{table}[h]
\caption{\label{nova2}Ranking of classifiers in DB10F considering the best configuration of parameters among the 1,000 random configurations.}
\centering
\begin{tabular}{@{}llcc}
\hline
\multirow{2}{*}{\#} & \multirow{2}{*}{\bf Classifier} & {\bf Average} & {\bf Deviation}  \\
&  & (\%) & (\%) \\
\hline
\textbf{1} & SVM & 98.8 & 0.7    \\
\textbf{2} & kNN & 94.3 & 1.8   \\
\textbf{3} & Random Forest & 88.7 & 1.9  \\
\textbf{4} & Logistic & 72.4 & 4.7  \\
\textbf{5} & C4.5& 67.1 & 2.8   \\
\textbf{6} & Simple Cart & 66.3 & 3.5  \\
\textbf{7} & Perceptron & 50.9 & 2.3 \\
\hline
\end{tabular}
\\ \ \\
\end{table}



\section*{Conclusions}

Machine learning methods have been applied to recognize patterns and classify instances in a wide variety of applications. Currently, several researchers/practioners with different expertise have employed computational tools such as Weka to study particular problems. Since the appropriate choice of parameters requires a certain knowledge of the underlying mechanisms behind the algorithms, oftentimes these methods are applied with their default configuration of parameters. Using the Weka software, we evaluated the performance of classifiers using distinct configurations of parameters in order to verify whether it is feasible to improve their performance. A summary of  the main results obtained in this study is provided in Table~\ref{t:tabela_resumo}.

\begin{table}[h]
\caption{\label{t:tabela_resumo}Summary of comparisons between classifiers considering two configuration of parameters (default or random) and two datasets (DB2$\mathbb{F}$ and DB10$\mathbb{F}$) we show the two classifiers achieving the best and worst performances.}
\centering
\begin{tabular}{@{}llll}
\hline
{\bf Type} & {\bf Case} & DB2$\mathbb{F}$ & DB10$\mathbb{F}$\\
\hline
\multirow{4}{*}{\bf Default} & \multirow{2}{*}{\bf Best}  & Naive Bayes & kNN        \\ \cline{3-4}
                         &                        & Logistic    & Perceptron \\ \cline{2-4}	
                         & \multirow{2}{*}{\bf Worst} & kNN         & C4.5       \\ \cline{3-4}
                         &                        & SVM         & Bayes Net  \\ \cline{2-4}	
\multirow{4}{*}{\bf Random} & \multirow{2}{*}{\bf Best}   & SVM & SVM        \\ \cline{3-4}
                         &                        & kNN & kNN \\ \cline{2-4}	
                         & \multirow{2}{*}{\bf Worst} & Simple Cart & SimpleCart       \\ \cline{3-4}
                         &                        & C4.5        & Perceptron  \\ \cline{2-4}				
\hline
\end{tabular}
\\ \ \\
\end{table}

The analysis of parameters in two-dimensional problems revealed that the Naive Bayes displays the best performance among all nine classifiers evaluated with default parameters. In this scenario, the SVM turned out to be the classifier with the poorest performance. When instances are described by a set of ten features, the kNN outperformed by a large margin the other classifiers, while the SVM retained its ordinary performance. When just one parameter is allowed to vary, there is not a large variation in the accuracy compared with the classification achieved with default parameters. The only exceptions are the parameter K of the kNN and parameter S of SVM (with Puk kernel). In these cases, the appropriate choice of the parameters enabled an average increase of 6\% in accuracy. Surprisingly, we found that when the same analysis is performed with a ten-dimensional dataset, the improvement in performance surpasses 20\% for the SVM. Finally, we developed a strategy in which all the configuration of parameters are chosen at random. Despite its outward simplicity, this strategy is useful to optimize SVM performance especially in high-dimensional problems, since the average increase provided by this strategy is higher than 20~\%.
%

Another important result arising from the experiments is the strong influence of the number of features on the performance of the classifiers. While small differences in performance across distinct classifiers were found in low-dimensional datasets, we found significative differences in performance when we analyzed problems involving several features. In high-dimensional tasks, kNN and SVM turned out to be the most accurate techniques when default and alternative parameters were considered, respectively.
Most importantly, we found that the behavior of the performance with the number of features follows three distinct patterns: (i) almost constant (Perceptron); (ii) monotonic increase (kNN), and (iii) monotonic decrease (Bayes net). These results suggest that number of features of the problem plays a key role on the choice of algorithms and, therefore, it should be considered in practical applications.

The results obtained here suggest that for low dimension classification tasks, Weka's default parameters provide accuracy rates close to the optimal value, with a few exceptions. The highest discrepancies arose in high-dimensional problems for the SVM, indicating that the use of default parameters in these conditions is not recommended in cases where the SVM must be employed. One could pursue this line of analysis further to probe the properties of classifiers with regard to other factors such as number of classes, number of instance per class and overlapping between classes. It is also very important to probe the performance in problems where the amount of instances employed to train is scarce, as it happens in occasions when data acquisition represents an expensive, painstaking endeavor.


\section*{Acknowledgments}
The authors acknowledge financial support from CNPq (Brazil) (grant numbers 573583/2008-0, 208449/2010-0, 308118/2010-3 and 305940/2010-4), FAPESP (grant numbers 2010/00927-9, 2010/19440-2, 2011/22639-8, 2011/50761-2, 2013/06717-4 and 2013/14984-2) and NAP eScience - PRP - USP.

\section*{References}

\newpage


\end{document}